\documentclass{style/hcrg-nasa}

\issuer{Harvard Computational Robotics Group}
\edition{Preprint}

\usepackage[utf8]{inputenc} 
\usepackage[T1]{fontenc}    
\usepackage{hyperref}       
\usepackage{url}            
\usepackage{booktabs}       
\usepackage{amsfonts}       
\usepackage{nicefrac}       
\usepackage{microtype}      
\usepackage{xcolor}         
\usepackage{colortbl}       
\usepackage{graphicx}
\usepackage{float}          
\usepackage{tikz}           
\usetikzlibrary{arrows.meta, backgrounds, calc}
\usepackage{xspace}
\usepackage{amsmath}

\newcommand{\our}[0]{\textsc{Ledger}\xspace}

\newenvironment{ack}{\section*{Acknowledgments}}{}

\title{Where Memory Belongs: \our{}, an Object Ledger for Memory-Augmented VLAs}
\author[1,2*]{Tanguy Dieudonné}
\author[1*]{Jack B. Jedlicki}
\author[1]{Heng Yang}
\affiliation[1]{Harvard University}
\affiliation[2]{ETH Zürich}
\contribution[*]{Equal contribution.\\
\texttt{tdieudonne@ethz.ch}\quad\texttt{jackbjed@g.harvard.edu}}

\abstract{
Memory is essential for long-horizon, partially observed robotic manipulation: a robot must remember which object was placed in a drawer, whose cup it moved, or how many action cycles have elapsed. Recent vision-language-action (VLA) models embed memory directly \textit{inside} the policy, but benchmarks show no single in-policy mechanism covers all spatio-temporal dimensions, trailing oracle methods by a wide margin. We argue that memory \emph{type} dictates where memory should reside: short-term perceptual memory (repetition, timing, retracing) belongs \textit{inside} the policy, while long-term object memory (persistent spatial state, containment, event history) belongs \textit{outside} as an explicit, readable record. We present \textbf{\our{}}, a harness that realizes this split over a single fine-tuned $\pi_{0.5}$ policy by pairing an in-policy frame-sampling memory with an external spatio-temporal object memory, the ledger (a SAM3 tracker, a VLM captioner of the demonstration, and an LLM that decides at step boundaries). On RoboMME, \our{} reaches the highest four-suite average among the evaluated methods, $64.3\%$ (vs. $45.9\%$ for the strongest prior method under identical evaluation), leading object reference ($60.7\%$ vs. $40.3\%$) and object permanence ($86.7\%$ vs. $56.2\%$) using a single set of weights. Choosing the memory source at runtime, from the instruction and the record, removes the need for a task-level router.
}

\begin{document}
\maketitle

\section{Introduction}

Memory allows humans to navigate and reason over partially-observed environments. For instance, if Alice wants to make a coffee latte using Bob's cup, she would have to remember where the coffee grounds are, the milk's expiry date, and where and which cup belongs to Bob. The information that decides the next action is no longer in the current frame. However, naively conditioning policies on long sequences of observations is computationally expensive, and policies start relying on spurious correlations in the history tokens instead of the immediate physical layout \citep{han2026geometricactionmodelrobot}. On RoboMME \citep{dai2026robomme}, a large-scale benchmark for memory-augmented robotic manipulation, the best in-policy memory reaches $44.5\%$ averaged over four memory suites as reported ($45.9\%$ under our reproduction in Sec.~\ref{sec:eval}), while the same backbone trained on the same demonstrations reaches $84.1\%$ when an oracle hands it a correctly resolved target. These results indicate that resolving what the policy has to act on is a major remaining bottleneck, rather than low-level execution alone.

We start from the observation that the tasks in such benchmarks demand two different kinds of memory. \emph{Short-term perceptual memory} is what the robot itself saw and did: how many swings it has completed, where the stick has already been, what the demonstrated motion looked like. This memory preserves rich low-level visual information that is difficult to capture with language, and is integrated inside the VLA. \emph{Long-term object memory} is what the world did: which object sits where now, which container was placed over which cube, which cube was picked earlier, and in what order objects were placed. This memory is persistent, outside the policy as an explicit record that can be read, checked, and reasoned over, and must survive occlusion and distractors. 

This raises a routing question: which memory should a given instruction use? Existing RoboMME methods do not ask it: each is one method run on every task suite, strong where its memory fits and weak elsewhere, and a learned router such as DIRECT \citep{dao2026directallocatetesttimecompute} picks the mechanism from per-task success labels. Instead, we want the routing decision to be made at runtime, from the instruction and the memory alone: at every step boundary a high-level policy reads the record and either emits a grounded target or hands control to the VLA's short-term memory through an empty subgoal, and a single set of $\pi_{0.5}$ \citep{black2025pi05} weights is fine-tuned so that both inputs remain in-distribution.

\begin{figure}[t!]
\centering
\resizebox{\textwidth}{!}{\usetikzlibrary{arrows.meta,backgrounds,calc} 

\definecolor{figslate}{HTML}{1E293B} 
\definecolor{figgray}{HTML}{64748B} 
\definecolor{figblue}{HTML}{3B82F6} 
\definecolor{figbluelight}{HTML}{EFF6FF} 
\definecolor{figpolicy}{HTML}{0D9488} 
\definecolor{figpolicylight}{HTML}{F0FDFA} 
\definecolor{figorange}{HTML}{D97706} 
\definecolor{figorangelight}{HTML}{FFFBEB} 
\definecolor{figpurple}{HTML}{7C3AED} 
\definecolor{figpurplelight}{HTML}{F5F3FF} 
\definecolor{figband}{HTML}{F8FAFC} 

\providecommand{\figh}{\sffamily\bfseries\small} 

\begin{tikzpicture}[ 
    x=1cm, y=1cm, font=\sffamily\footnotesize, 
    >={Latex[length=2.2mm,width=1.6mm]}, 
    flow/.style={-Latex, draw=figgray, line width=0.8pt}, 
    flowB/.style={-Latex, draw=figblue, line width=0.9pt}, 
    flowG/.style={-Latex, draw=figpolicy, line width=0.9pt}, 
    flowO/.style={-Latex, draw=figorange, line width=1.0pt}, 
    flowP/.style={-Latex, draw=figpurple, line width=0.9pt, dashed, dash pattern=on 3.5pt off 2.5pt}, 
    basebox/.style={draw=black!25, fill=white, rounded corners=6pt, line width=0.8pt, align=center, inner xsep=10pt, inner ysep=7pt}, 
    inputbox/.style={basebox, draw=figgray!50, fill=black!2, line width=1pt}, 
    bluebox/.style={basebox, draw=figblue, fill=figbluelight, line width=1pt}, 
    policybox/.style={basebox, draw=figpolicy, fill=figpolicylight, line width=1pt}, 
    orangebox/.style={basebox, draw=figorange, fill=figorangelight, line width=1pt}, 
    purplebox/.style={basebox, draw=figpurple, fill=figpurplelight, line width=1pt}, 
    lab/.style={font=\sffamily\itshape\tiny, text=figgray, align=center, inner sep=2pt}, 
    labB/.style={font=\sffamily\itshape\tiny, text=figblue, align=center, inner sep=2pt}, 
    labG/.style={font=\sffamily\itshape\tiny, text=figpolicy, align=center, inner sep=2pt}, 
    labO/.style={font=\sffamily\itshape\tiny, text=figorange, align=center, inner sep=2pt}, 
    labP/.style={font=\sffamily\itshape\tiny, text=figpurple, align=center, inner sep=2pt}, 
    title/.style={anchor=west, font=\sffamily\bfseries\small, text=figslate}, 
] 

  \begin{scope}[on background layer] 
    \filldraw[fill=figband, draw=black!10, rounded corners=8pt, line width=0.8pt] (3.50,8.60) rectangle (19.40,12.10); 
    \filldraw[fill=figband, draw=black!10, rounded corners=8pt, line width=0.8pt] (3.50,4.80) rectangle (19.40,8.10); 
    \filldraw[fill=figband, draw=black!10, rounded corners=8pt, line width=0.8pt] (3.50,0.80) rectangle (19.40,4.30); 
  \end{scope} 

  \node[title] at (3.75,11.75) {Long-Term Object Ledger}; 
  \node[title] at (3.75,7.75)  {One Decision Per Step Boundary}; 
  \node[title] at (3.75,1.15)  {Short-Term Perceptual Memory}; 

  \node[inputbox, minimum width=2.4cm] (obs) at (1.50,10.22) {{\figh\color{figgray} Observations}\\[1mm]\color{figgray}\itshape\tiny demo video $\cdot$ camera frame}; 
  \node[inputbox, minimum width=2.2cm] (instr) at (5.00,6.45) {{\figh\color{figgray} Instruction}}; 

  \node[bluebox, minimum width=3.5cm] (tracker) at (5.50,10.90) {{\figh\color{figblue}SAM3 Tracker}\\[0.5mm]\color{figblue}\itshape\scriptsize object positions, every step}; 
  \node[bluebox, minimum width=3.5cm] (narr)    at (5.50,9.55)  {{\figh\color{figblue}Video-VLM Narration}\\[0.5mm]\color{figblue}\itshape\scriptsize chronological actions}; 

  \node[bluebox, align=left] (record) at (11.25,10.22) 
    {{\figh\color{figblue}Object Record}\\[1mm] 
     {\color{figblue}\ttfamily\tiny B: green cube, position [70,163]}\\ 
     {\color{figblue}\ttfamily\tiny \phantom{B:} rests [1,61]@[52,140] [97,118]@[70,163]}\\ 
     {\color{figblue}\ttfamily\tiny \phantom{B:} covered\_since 118}\\ 
     {\color{figblue}\ttfamily\tiny demonstration: "picks up B, places it on A"}}; 

  \node[bluebox, minimum width=3.2cm] (occl)  at (17.30,10.90) {{\figh\color{figblue}Occlusion Query}\\[0.5mm]\color{figblue}\itshape\scriptsize arm vs. cover}; 
  \node[bluebox, minimum width=3.2cm] (frame) at (17.30,9.55)  {{\figh\color{figblue}Annotated Frame}\\[0.5mm]\color{figblue}\itshape\scriptsize positions drawn}; 

  \node[orangebox, minimum width=6.8cm] (planner) at (11.25,6.45) 
    {{\figh\color{figorange}Planner (LLM)}\\[1mm]\color{figorange}\itshape\scriptsize reads record + frame + history $\rightarrow$ generates step subgoal}; 

  \node[purplebox, minimum width=3.2cm] (log) at (17.30,6.45) {{\figh\color{figpurple}Episode Log}\\[0.5mm]\color{figpurple}\itshape\scriptsize steps $\cdot$ gripper $\cdot$ plan}; 

  \node[policybox, minimum width=3.8cm] (pi) at (5.50,2.55) {{\figh\color{figpolicy}Policy $\boldsymbol{\pi_{0.5}}$}\\[1mm]\color{figpolicy}\itshape\scriptsize frame-sampling memory}; 

  \node[inputbox, minimum width=2.2cm] (robot) at (11.25,2.55) {{\figh\color{figgray} Robot}\\[0.5mm]\color{figgray}\itshape\scriptsize joint state}; 
  \node[purplebox, minimum width=3.2cm] (prop) at (17.30,2.55) {{\figh\color{figpurple}Proprioception}\\[0.5mm]\color{figpurple}\itshape\scriptsize grasp / release / press}; 


  \draw[flow] ($(obs.east)+(0,0.35)$) -- (tracker.west); 
  \draw[flow] ($(obs.east)+(0,-0.35)$) -- (narr.west); 
  \draw[flow] (obs.south) |- node[lab, pos=0.75, above] {frames} (pi.west); 

  \draw[flowB] (tracker.east) -- (record.west |- tracker.east); 
  \draw[flowB] (narr.east) -- (record.west |- narr.east); 
  \draw[flowB, Latex-Latex] (record.east |- occl.west) -- node[labB, above] {lost?} (occl.west); 
  \draw[flowB] (record.east |- frame.west) -- (frame.west); 

  \draw[flowB] (record.south) -- node[labB, right=2pt] {record} (planner.north); 
  \draw[flowB] (frame.south) -- ++(0,-0.4) -| node[labB, pos=0.2, below=2pt] {frame} ($(planner.north east)+(-0.8,0)$); 
  \draw[flow] (instr.east) -- (planner.west); 

  \draw[flowO] ($(planner.south west)+(0.8,0)$) -- node[labO, right=4pt, pos=0.4, align=left] {subgoal: e.g., \texttt{pick up cube at <70,163>}\\ \emph{or} empty} ++(0,-1.4) -| (pi.north); 

  \draw[flowG] (pi.east) -- node[labG, above=1pt] {actions} (robot.west); 
  \draw[flow] (robot.east) -- node[lab, above=1pt] {state} (prop.west); 
  \draw[flowP] (prop.north) -- node[labP, right=2pt] {step done} (log.south); 
  \draw[flowP] (log.west) -- node[labP, above=2pt, pos=0.5] {done so far} (planner.east); 

\end{tikzpicture}}
\caption{\textbf{\our{} Overview}. Two memories serve one $\pi_{0.5}$ policy. \textbf{Outside the policy}, a SAM3 tracker seeded from the first frame records where each object rests and when it is covered, and a video VLM transcribes the demonstration when there is one. \textbf{Inside the policy}, a frame-sampling modulator over past observations carries repetition, timing, and motion. At each step boundary the LLM planner reads the instruction, the record, an annotated frame, and the completed steps, and either grounds a coordinate target or defers to the policy's own memory with an empty subgoal.}
\label{fig:overview}
\end{figure}

Overall, our contributions are:
\begin{itemize}
    \item \textbf{A principle for where memory belongs.} Persistent world and event state (\emph{what the world did}) is kept outside the policy as an explicit spatio-temporal record, whereas short-term perceptual state (\emph{what the robot just observed and executed}) remains inside the policy, as history-conditioned features. Across RoboMME, performance on every task suite is led by the specific memory source predicted by this principle.
    \item \textbf{\our{}.} An explicit object-centric memory that operationalizes this principle by combining per-object tracker-derived stationary intervals and occlusion events, VLM-generated demonstration transcripts, and a geometric co-location rule that unifies both representations for high-level planning. A single $\pi_{0.5}$ policy acts either on a target grounded in the ledger or from its own frame memory, and the planner chooses between them at runtime, without a task-level router.
\end{itemize}
On RoboMME, \our{} reaches the highest average success among the evaluated methods, $64.3\%$ against $45.9\%$ for the strongest prior baseline, and outperforms all prior methods on object permanence and reference.

\section{Related Work}

\paragraph{What manipulation must remember.}
RoboMME \citep{dai2026robomme} scripts sixteen tasks in four suites on one Franka arm, one demonstration set, and fine-tunes the same $\pi_{0.5}$ backbone, reaching $84.1\%$ with oracle grounded subgoals, $32.7\%$ when a fine-tuned VLM writes the subgoals from the current image, and $44.5\%$ with its best in-policy memory, from which the authors conclude that memory representations are complementary rather than exclusive. LIBERO-Mem \citep{chung2025liberomem} makes the object-centric case with visually identical bowls that only their interaction history tells apart; even its own slot-based policy, given oracle per-object subgoals, completes only $14.8\%$ of subgoals, and $\pi_0$ completes $5\%$. RMBench \citep{chen2026rmbench} defines memory complexity as the number of past observations that an optimal policy must retain. Its Mem-0 baseline improves $\pi_{0.5}$ from $10.4\%$ to $42.0\%$ using a planner invoked at subtask boundaries over a memory of completed subtasks, with groundtruth signals worth $17$ points more. RoboMemArena \citep{robomemarena2025} tests horizons of over a thousand steps, where frozen VLM planners collapse to $8.7\%$. Together, these benchmarks fix \textit{what} must be remembered and the gaps, but leave open \textit{how} memory should be represented, \textit{where} it should live, and \textit{which} representation serves \textit{which} kind of memory. We use RoboMME because it spans four memory types with one embodiment and one training set.

\paragraph{Memory inside the policy.}
The in-policy memory variants from RoboMME compress the observation history in different ways. Uniform frame sampling can modulate the action expert, token dropping removes temporal redundancy in image patches, and recurrent memory compresses the visual token history sequence into fixed-size latent states carried through recurrent updates \citep{bulatov2022recurrent}. HAMLET \citep{koo2026hamlet} compactly encodes perceptual information at each step into moment tokens, integrated as memory features used during action prediction, and Gated Memory Policy \citep{gao2026gatedmemorypolicy} employs a learned memory gate mechanism that decides when to cross-attend a fixed window. SAM2Act+ \citep{fang2025samact} stores the policy's own past action heatmaps in a fixed queue and cannot retain semantic information such as color. EchoVLA \citep{lin2026echovlasynergisticdeclarativememory} fuses a voxel scene memory and a token queue into mobile-manipulation diffusion heads and degrades under dynamic occlusions. $\pi_{0.6}$-MEM \citep{pimem} and $\pi_{0.7}$ \citep{pi07} combine a short and long-term memory, using compressed video and text representations. MemER \citep{sridhar2026memer}, the strongest prior method on permanence and reference, proposes a hierarchical policy framework where a high-level policy uses selected keyframes and the most recent frames when producing text instructions for a low-level policy to execute; it falls on RoboMME to $38\%$ and $21\%$ on the swap tasks where oracle grounding reaches $99\%$ and $80\%$. These models remember what the robot saw, but none keeps, per object, where it rested and what came to cover it.

\paragraph{Explicit records and harnesses.} Object maps and scene graphs \citep{gu2024conceptgraphs, hughes2022hydra} provide an interface for inference and reasoning, but only keep current information: ConceptGraphs relocates a moved object by asking an LLM for plausible containers, not by consulting where it last rested. Compose by Focus \citep{qi2026compose} conditions diffusion-based atomic skills on task-relevant 3D scene graphs and composes them with a VLM planner, improving robustness to distractors and novel scene compositions. Systems that place a reasoning model over a fixed policy expose state the same way. THEA \citep{thea2026} keeps a persistent symbolic scene graph, re-renders it into the planner's context at every turn, and admits execution-derived edits only once an evaluator has confirmed a tool's post-condition, yet the graph is a present-tense snapshot with a freshness stamp: it records neither where an object rested earlier nor what came to cover it. Pigey \citep{pigey} lifts a frozen $\pi_{0.5}$ from $12.8\%$ to $53.3\%$ on LIBERO-PRO with a perceive-act-verify loop whose memory is the history of attempted subgoals, their outcomes, and whose verification is a gripper sensor plus the planning VLM itself. Robot Critics \citep{sudhakar2026robotcriticssweatsmall} finds frontier VLMs near chance ($54.7\%$ against $50\%$) at fine-grained success judgment. Finally, which memory a task needs has so far been decided before execution. DIRECT \citep{dao2026directallocatetesttimecompute} trains a router on the per-task success and cost of a fixed pool of RoboMME memories and uses it once per episode from the initial observation and instruction. \our{} keeps the record these systems lack: what happened in the scene as stationary intervals and event descriptions per object, confirming step boundaries from proprioception rather than a VLM verdict, and letting the planner decide, from the instruction and the ledger, whether to ground execution.

\section{\our}

\subsection{Preliminaries}

\paragraph{Problem Setting.} We consider language-conditioned manipulation policies trained to model the conditional distribution $\pi(\mathbf{A_t}|o_t, l_t)$, where $\mathbf{A_t} = [a_t, a_{t+1}, \cdots, a_{t+H-1}]$ is an action chunk modeled from the current timestep $t$ up to $H$ timesteps, $o_t$ is the robot's current sensor observation, and $l_t$ is the language instruction. At each timestep $t$, the robot receives a multi-view RGB observation $\mathbf{I_t} = [I_t^1, I_t^2, \cdots, I_t^n]$, and $\mathbf{q_t}$ are the proprioceptive inputs from the robot (joint angles and gripper state), hence $o_t = [\mathbf{I_t}, \mathbf{q_t}]$. The demonstration of a task, the covering of an object, or the number of repetitions completed so far are not recoverable from $o_t$ alone.

In order to execute complex, long-horizon tasks, we follow the hierarchical factorization \citep{shi2025hi} of the robot policy into a low-level control policy $\pi_{\text{LL}}$ and a high-level policy $\pi_{\text{HL}}$. The low-level policy only carries a short-term perceptual memory (Sec.~\ref{sec:frame}), while the high-level policy carries the long-term object-centric memory $m_t$ (Sec.~\ref{sec:objmem}) and decides what the low-level policy should act on by giving it language subgoals (Sec.~\ref{sec:loop}). We decompose action prediction as
\[
\pi (a_{t:t+H}, l_{t+1}, m_{t+1} | o_{t-T:t}, m_t, g) \approx \pi_{\text{LL}}(a_{t:t+H}|o_{t-K:t}, l_{t+1}, g)\; \pi_{\text{HL}}(l_{t+1}, m_{t+1} | o_t, m_t, g),
\]
where $g$ is the task instruction, $l_{t+1}$ is the subgoal handed to the low-level policy (possibly empty), and $K \ll T$.

\subsection{Long-Term Memory: Object-Centric Spatio-Temporal}
\label{sec:objmem}

Long-term memory enters only through the high-level policy's language input, as a record we call the \emph{ledger}. 
Let $\mathcal{O}$ be the set of objects enumerated in the initial observation, and for each $i \in \mathcal{O}$, let $p_i(t) \in \mathbb{R}^2 \cup \{\bot\}$ denote its tracked image coordinate at timestep $t$ ($\bot$ while its track is lost) across both the optional task demonstration and the execution rollout. The ledger at time $t$ is formulated as
\begin{equation}
\mathcal{L}_t = \big( \{ (d_i,\; p_i(t),\; \mathcal{R}_i(t),\; c_i) \}_{i \in \mathcal{O}},\; \mathcal{N} \big),
\label{eq:ledger}
\end{equation}
where $d_i$ is the textual object description, $c_i$ is the timestamp at which object $i$ was covered by an entity other than the robot arm (or $\emptyset$), $\mathcal{N}$ is the action transcript of the demonstration (or $\emptyset$), and $\mathcal{R}_i(t)$ is the list of \emph{stationary intervals} of object $i$: the maximal time windows over which its track stays still, each stored with its start, its end, and its mean position $\bar p_i$. Spatial reasoning then reduces to one geometric primitive, co-location: two positions are the same place when they lie closer than a tolerance of under one object width, and every spatial query the planner evaluates is derived from it. Object $j$ was placed onto $i$ if a stationary interval of $j$ that follows a placement is co-located with $\bar p_i$, the number of such intervals is the number of placements, and object $i$ is currently covered by $k$ if $c_i$ is set and $p_k(t)$ is co-located with $p_i(t)$.

\paragraph{Building the ledger.} At the start of an episode, the planner model enumerates the visible objects in the initial frame. A SAM3 tracker \citep{carion2026sam} then tracks each instance across both the demonstration video (when available) and the live execution rollout. Textual descriptions, coordinates, visibility flags, and stationary intervals are computed directly from these trajectories. Cover timestamps $c_i$ are maintained by the tracker's recovery gate: whenever a trajectory is interrupted, the planner model is shown the annotated frame and asked whether the object is hidden by the robot arm or by another entity. Occlusions caused by the arm trigger a wait state, whereas environmental occlusions record the timestamp $c_i$. The recovered trajectory subsequently follows the covering entity, ensuring the coordinate of a covered object continuously tracks the location of its cover. This mechanism explicitly captures spatial state modifications, such as container swaps, that static, single-frame scene graphs cannot represent. Appendix~\ref{app:ledger} illustrates an instance of the ledger provided to the high-level planner.

\paragraph{The action transcript.} Pure geometric tracking cannot determine action semantics: which object was grasped, placed, or actuated, and in what sequence. When an episode includes a demonstration video, a video VLM (Qwen3-VL-30B) analyzes frames annotated with tracked instance identifiers to produce an ordered natural-language account of the robot's execution alongside press-timing intervals. This action transcript is generated once per episode. For tasks without a demonstration video, the transcript is marked empty and the planner relies entirely on real-time observations. Empirically, the transcript reliably captures event sequences, temporal ordering, and repetition counts, but is less accurate at disambiguating destinations among visually identical objects; Appendix~\ref{app:ledger} shows an instance. The high-level planner is therefore instructed to extract event structure from the action transcript while resolving all spatial target coordinates via stationary-interval co-location.

\subsection{Short-Term Memory: Frame Sampling as Feature Modulation}
\label{sec:frame}

The low-level control policy is based on $\pi_{0.5}$ \citep{black2025pi05}, where a vision-language backbone encodes input image observations and language prompts into visual-textual tokens $\mathbf{u}_t$. A flow-matching action expert then denoises a noisy action chunk $\mathbf{s}^0_t$ across $L$ residual layers (omitting explicit denoising timestep conditioning for brevity):
\begin{equation}
\tilde{\mathbf{s}}^k_t = \mathbf{s}^{k-1}_t + \mathrm{Attn}^k\big(\mathbf{s}^{k-1}_t, \mathbf{u}_t\big), \qquad
\hat{\mathbf{s}}^k_t = \gamma^k_t \odot \mathrm{Norm}(\tilde{\mathbf{s}}^k_t) + \beta^k_t, \qquad
\mathbf{s}^k_t = \tilde{\mathbf{s}}^k_t + \mathrm{MLP}^k\big(\hat{\mathbf{s}}^k_t\big),
\label{eq:expert}
\end{equation}
where $(\gamma^k_t, \beta^k_t) = (1, 0)$ in the base, unmodulated model. 

Short-term perceptual memory is integrated directly into this action expert. Following \citet{dai2026robomme}, the policy retains $N=32$ uniformly sampled past observations $\mathcal{S}_t$. The vision encoder $\phi$ extracts features from each historical frame, which are then pooled into a compact token set $\mathbf{M}_t = [\mathrm{Pool}(\phi(o_\tau))]_{\tau \in \mathcal{S}_t}$. Every expert layer queries these visual tokens via cross-attention to compute feature modulation parameters:
\begin{equation}
(\gamma^k_t, \beta^k_t) = (1, 0) + f^k\big(\mathrm{Attn}^k_{\mathrm{mod}}(\tilde{\mathbf{s}}^k_t, \mathbf{M}_t)\big),
\label{eq:modulator}
\end{equation}
where $f^k$ is initialized close to zero, so that fine-tuning starts from the unmodulated policy. This in-policy memory provides an explicit execution trace of recent robot observations and joint actions, such as swing counts, trajectory histories, and motion patterns. However, it cannot maintain persistent object identities across severe occlusions or keep an explicit record of where each object was moved.

Long-term memory, by contrast, is supplied to the same low-level policy exclusively through the language channel as a grounded subgoal prompt. To enable a single unified policy to operate across both memory modalities, we introduce a \emph{task-conditional subgoal dropout curriculum} during fine-tuning. On temporal and procedural benchmark suites, subgoals are dropped during training with probability $0.9$, forcing the policy to execute tasks relying solely on its internal perceptual memory. On spatial and object suites, subgoals are dropped with probability $0.15$, preserving the policy's capacity to leverage external spatial targets for occluded objects.

Both prompt conditions, with and without a grounded subgoal, are thus in-distribution at deployment, and the planner selects between them at runtime. The suite labels serve only in training; deployment receives no task or suite identifier. 

Without this curriculum the subgoal is a training shortcut on procedural tasks: it prescribes the motion step by step (``move backward, move right''), so the policy reads the answer from text and never builds the perceptual memory it must use once the subgoal is withheld.

\subsection{Deciding at Step Boundaries}
\label{sec:loop}

The high-level policy is an LLM planner (Claude Sonnet 5) that operates over the ledger and one annotated frame per decision, never the video stream or the motor commands. It is invoked at the start of an episode and whenever an execution step completes. Each query supplies the planner with the task instruction, the current ledger, the action transcript (if available), an annotated observation frame marking tracked objects with identifier letters, and an execution log listing completed steps alongside their completion timestamps. The planner returns a structured plan consisting of a sequence of subgoals, where each subgoal uses a sentence template from the low-level policy's training vocabulary with a slot for a spatial coordinate target. Only the immediate first subgoal is executed, and subsequent decisions are deferred until the low-level policy completes the step.

Replies are parsed strictly: a plan must draw its sentences from the low-level vocabulary, alternate picks and places so that the gripper never holds two objects, and reference only tracked identifiers. An invalid reply is re-prompted with the validation error, and an episode that keeps failing validation is recorded as a failure rather than executed on a guess. 

The planner may also emit an \emph{empty} subgoal, which hands the rest of the episode to the low-level policy's own memory (Sec.~\ref{sec:frame}); grounding or deferring is thus an explicit planner action, taken from the instruction and the ledger.

\paragraph{Why step boundaries.} A static plan generated at episode onset cannot adapt to downstream events, such as an object becoming visible only after a button press or containers swapping during manipulation. Conversely, revising plans at fixed time intervals or on visual changes induces instabilities caused by scene changes during execution. Evaluating decisions strictly at step completion mitigates both issues: replanning occurs at discrete execution boundaries when the gripper reaches a known, stable state, avoiding mid-motion intervention.

\paragraph{Step completion from proprioception.} Step completion is determined entirely from robot proprioception rather than visual trackers or VLM evaluators. The gripper aperture is categorized into three discrete regimes—shut, hold, and open—normalized relative to the maximum observed aperture. A grasp corresponds to entering the hold regime, a release to transitioning from hold to open, and a press to a reach and retreat of the arm with the gripper shut. Regime transitions must persist across consecutive control steps to prevent false triggers, with extended temporal filtering applied to the hold state to ignore transient aperture states during full gripper closure. Each planned step specifies its terminating event type, and live target coordinates are queried from the tracker at the exact onset of step execution.

\section{Experimental Evaluation}
\label{sec:eval}

\paragraph{Experimental protocol.} All methods are evaluated on the official RoboMME test split across 16 tasks, 50 evaluation episodes per task, and three random seeds. Results are reported using the final model checkpoint for each method: our fine-tuned policy at $80\mathrm{k}$ steps, alongside the released final checkpoints for FrameSamp+Modul and MemER. Success rates represent mean performance across all evaluation seeds. 

In addition to prior baselines, Table~\ref{tab:main} includes a \emph{tracker-only} ablation variant of our harness. On spatial and object tasks (Permanence and Reference), a fine-tuned VLM selects among SAM3 tracks without access to historical event records; on temporal and procedural tasks (Counting and Imitation), the low-level $\pi_{0.5}$ policy executes directly using its internal short-term memory without subgoals. This baseline explicitly isolates the gains provided by persistent geometric tracking prior to integrating historical event memory.

\subsection{One Policy, Best Across Memory Types}
\label{sec:eval-perm}

Table~\ref{tab:main} reports success rates across all four RoboMME memory benchmark suites. Prior methods that demonstrate strength on a single suite exhibit severe performance degradation on others: under our evaluation protocol, FrameSamp+Modul achieves $69.2\%$, $25.7\%$, $37.3\%$, and $51.3\%$ on Counting, Permanence, Reference, and Imitation, respectively, whereas MemER obtains $49.2\%$, $56.2\%$, $40.3\%$, and $27.3\%$. 

In contrast, \our{} uses a unified set of weights and a single high-level planner across all four suites. On Counting and Imitation, \our{} trails the suite-specific leader, FrameSamp+Modul, by two points ($66.8\%$ vs. $69.2\%$) and by eight points ($43.0\%$ vs. $51.3\%$), with both methods running on the same in-policy perceptual memory.

On Permanence and Reference, external object memory provides the largest gains: Permanence reaches $86.7\%$ (vs. $56.2\%$ for MemER), with the first three tasks ranging from $90.0\%$ to $92.7\%$ and ButtonUnmaskSwap reaching $72.7\%$. Reference achieves $60.7\%$ (vs. $40.3\%$), with individual task success rates spanning $55.3\%$ to $64.0\%$. Across the three seeds, the standard deviation of \our{}'s suite means is $2.8$, $3.5$, $5.2$, and $2.2$ points on Counting, Permanence, Reference, and Imitation, and $0.9$ points on the 16-task average (Appendix~\ref{app:variance}).

\newcommand{\best}[1]{\textcolor{red}{#1}}
\begin{table}[t]
\centering
\caption{\textbf{Main results.} RoboMME success rate (\%) on the test split, 50 episodes per task, mean over evaluation seeds; Avg is the mean over the 16 tasks. Upper block: as reported by \citet{dai2026robomme}. Lower block: methods run by us under the protocol of Sec.~\ref{sec:eval}. Oracle and QwenVL are GroundSG with ground-truth subgoals and with the benchmark's fine-tuned Qwen3-VL subgoal generator; FrameSamp is FrameSamp+Modul; Tracker is the tracker-only variant of our harness. \best{Red} marks the best method within each block; Human and Oracle are reference rows. Seed spreads are in Appendix~\ref{app:variance}.}
\label{tab:main}
\setlength{\tabcolsep}{1.2pt}
\scriptsize
\resizebox{\textwidth}{!}{%
\begin{tabular}{@{}l|cccc|cccc|cccc|cccc|c@{}}
\toprule
\textbf{Method} & \multicolumn{4}{c|}{\textbf{Counting}} & \multicolumn{4}{c|}{\textbf{Permanence}} & \multicolumn{4}{c|}{\textbf{Reference}} & \multicolumn{4}{c|}{\textbf{Imitation}} & \textbf{Avg} \\
 & \shortstack{Bin\\Fill} & \shortstack{Pick\\Xtimes} & \shortstack{Swing\\Xtimes} & \shortstack{Stop\\Cube} & \shortstack{Video\\Umsk} & \shortstack{Button\\Umsk} & \shortstack{Video\\UmskS} & \shortstack{Button\\UmskS} & \shortstack{Pick\\HighL} & \shortstack{Video\\Repick} & \shortstack{Video\\PlcBtn} & \shortstack{Video\\PlcOrd} & \shortstack{Move\\Cube} & \shortstack{Insert\\Peg} & \shortstack{Pattern\\Lock} & \shortstack{Route\\Stick} & \\
\midrule
\multicolumn{18}{@{}l}{\textit{As reported by \citet{dai2026robomme}: three seeds $\times$ three checkpoints}} \\
Human & 96.0 & 100.0 & 80.0 & 78.0 & 90.0 & 92.0 & 92.0 & 90.0 & 92.0 & 92.0 & 98.0 & 90.0 & 90.0 & 98.0 & 84.0 & 86.0 & 90.5 \\
Oracle & 85.8 & 100.0 & 100.0 & 49.7 & 98.8 & 95.0 & 99.2 & 80.2 & 83.3 & 97.3 & 100.0 & 100.0 & 87.8 & 15.6 & 97.0 & 55.6 & 84.1 \\
QwenVL & 52.0 & \best{92.7} & 7.3 & 0.0 & \best{88.7} & 24.0 & 30.7 & 14.0 & 15.1 & 25.3 & 54.0 & 31.8 & 71.6 & 3.3 & 6.7 & 6.0 & 32.7 \\
FrameSamp & 39.6 & 87.3 & \best{92.0} & \best{42.0} & 32.7 & 25.1 & 24.4 & 18.2 & 22.9 & \best{30.4} & \best{60.0} & \best{32.0} & 77.8 & \best{7.6} & \best{53.6} & \best{66.7} & \best{44.5} \\
MemER & \best{56.7} & 79.3 & 59.3 & 0.0 & 81.3 & \best{72.0} & \best{38.0} & \best{21.3} & \best{70.7} & 25.3 & 30.0 & 26.0 & \best{82.7} & 6.7 & 16.7 & 12.0 & 42.4 \\
\midrule
\multicolumn{18}{@{}l}{\textit{Reproduced by us: released final checkpoint, three seeds, one protocol}} \\
FrameSamp & 44.0 & 89.3 & \best{96.0} & \best{47.3} & 34.0 & 26.7 & 24.0 & 18.0 & 20.0 & 33.3 & 58.0 & 38.0 & \best{81.3} & 5.3 & \best{55.3} & 63.3 & 45.9 \\
MemER & \best{55.3} & 78.0 & 61.3 & 2.0 & 84.7 & 82.0 & 36.0 & 22.0 & \best{73.3} & 26.0 & 30.7 & 31.3 & 80.7 & 3.3 & 16.7 & 8.7 & 43.3 \\
Tracker & 41.3 & 90.0 & 92.7 & 40.0 & \best{92.7} & 88.0 & \best{92.0} & \best{76.0} & 18.7 & 28.0 & 53.3 & 30.7 & 76.0 & 7.3 & 41.3 & \best{64.0} & 58.3 \\
\textbf{\our{}} & 50.0 & \best{92.7} & 88.0 & 36.7 & \best{92.7} & \best{90.0} & 91.3 & 72.7 & 55.3 & \best{60.7} & \best{62.7} & \best{64.0} & 60.7 & \best{10.7} & 42.0 & 58.7 & \best{64.3} \\
\bottomrule
\end{tabular}}
\end{table}

\subsection{The Optimal Memory Source is Task-Dependent}
\label{sec:eval-ablation}

Table~\ref{tab:main} also says \emph{where} memory should reside. Read by information source, its lower block compares four: FrameSamp acts from in-policy frame memory with no subgoal, Tracker grounds a subgoal from the current tracks with no record of the past, \our{} grounds it from the ledger, and the Oracle row is the ceiling of a ground-truth subgoal. The pattern is consistent across suites:
\begin{itemize}
    \item \textbf{Spatial (Permanence):} a tracked target is enough. Tracker and \our{} reach $87.2\%$ and $86.7\%$ against $25.7\%$ without a subgoal and a $93.3\%$ oracle, and both score at least $88\%$ on the first three Permanence tasks. Tracking resolves a persistent target, so spatial memory belongs \emph{outside} the policy.
    \item \textbf{Temporal (Counting) and Procedural (Imitation):} the policy's own memory is the best source. FrameSamp, which never receives a subgoal, leads both suites ($69.2\%$ and $51.3\%$), while the benchmark's subgoal-driven pipeline trails far behind ($38.0\%$ and $21.9\%$ for QwenVL). A grounded target is not what these tasks lack: on Counting it neither helps nor hurts ($66.8\%$ for \our{} against $66.0\%$ for the same policy without subgoals), and on MoveCube, where our planner grounds a target instead of deferring, it costs fifteen points ($60.7\%$ vs. $76.0\%$). Repetitions, timing, and motion patterns are not a coordinate, so temporal and procedural memory belong \emph{inside} the policy.
    \item \textbf{Object (Reference):} neither source suffices alone. Tracks with no record of the past give $32.7\%$ and in-policy memory $37.3\%$, against a $95.2\%$ oracle; the ledger lifts the suite to $60.7\%$. The policy can execute the skill given the right target, but identifying that target requires a record of past events, which is what Sec.~\ref{sec:objmem} supplies.
\end{itemize}
Given a correctly resolved target, the benchmark's oracle-conditioned policy executes well on every suite (Oracle: $83.9\%$, $93.3\%$, $95.2\%$, and $64.0\%$), so what remains is \emph{resolving} what to act on and when, which \our{} does according to memory type.

\paragraph{What the decision loop contributes.} Table~\ref{tab:objmem-ablation} varies the decision loop of Sec.~\ref{sec:loop} over the ledger of Sec.~\ref{sec:objmem} on the five demonstration tasks. With a single plan made at the start of the episode, the planner resolves persistent targets (Unmask $90\%$) but cannot react to what happens during execution. Re-deciding from scratch after every completed step, with the cover state exposed, lifts the two tasks whose key event occurs after execution starts, PlaceButton and PlaceOrder ($58\% \rightarrow 66\%$ and $60\% \rightarrow 66\%$), but destabilizes Repick ($60\% \rightarrow 28\%$): without a record of its own earlier decisions, the planner re-attributes completed repetitions to the remaining goal and miscounts. Carrying the remaining plan across decisions and counting repetitions from the execution log restores Repick to $54\%$ and keeps PlaceButton at $66\%$. On the container-swap task UnmaskSwap, re-deciding lifts success from $68\%$ to $86\%$ and carrying the plan to $92\%$.

\begin{table}[t]
\centering
\caption{\textbf{Ablation of the decision loop on the demonstration-guided tasks} (success \%, one evaluation seed, 50 episodes per cell, checkpoint $80$k; a cell's binomial standard error is up to $7$ points). The last row is the configuration of Table~\ref{tab:main}, which averages three seeds. Re-deciding at step boundaries lets the planner react to events that occur during execution; carrying the remaining plan across decisions keeps the count of completed repetitions.}
\label{tab:objmem-ablation}
\small
\setlength{\tabcolsep}{4pt}
\begin{tabular}{lccccc}
\toprule
Decision loop over the ledger & Unmask & UnmaskSwap & Repick & PlaceButton & PlaceOrder \\
\midrule
Plan once at the start of the episode & 90 & 68 & 60 & 58 & 60 \\
+ cover state, re-decide after each step & 92 & 86 & 28 & 66 & 66 \\
+ remaining plan carried across decisions & 92 & 92 & 54 & 66 & 56 \\
\bottomrule
\end{tabular}
\end{table}

\paragraph{Can the high-level policy choose the memory source at runtime?} Whether to ground a target or to defer to the policy's own memory is a planner action, taken from the instruction and the ledger and never from task or suite metadata. We evaluate it in the complete closed loop, with the planner deployed on all 16 tasks. Table~\ref{tab:main} reports this always-on configuration, and on Counting and Imitation its Tracker row is the same policy in memory mode, so the comparison reads directly off the table. The planner defers on every \emph{PatternLock} and \emph{RouteStick} episode and on two thirds of \emph{StopCube} episodes, where deferring succeeds more often than grounding, and stays within six points of memory mode on all three ($42.0\%$ vs. $41.3\%$, $58.7\%$ vs. $64.0\%$, and $36.7\%$ vs. $40.0\%$). Where the instruction names an object to act on, it grounds instead: on \emph{BinFill} and \emph{PickXtimes} it plans the picks, the places, and the final press, counting them from confirmed step completions, and improves on memory mode ($50.0\%$ vs. $41.3\%$ and $92.7\%$ vs. $90.0\%$), so that Counting as a whole matches memory mode ($66.8\%$ vs. $66.0\%$). The deficit on Imitation ($43.0\%$ vs. $47.2\%$) stems mostly from \emph{MoveCube}, where the planner grounds a coordinate target for a motion that should be reproduced from memory ($60.7\%$ vs. $76.0\%$). On the benchmark's instructions, the planner thus infers the memory allocation at runtime from the instruction and the ledger, at the cost of one task.

\section{Discussion and Future Work}
\label{sec:discussion}

While \our{} leads the evaluated methods across diverse memory modalities, several limitations remain. First, our implementation operates in simulation, grounding targets in 2D image coordinates. Extending this architecture to mobile manipulation will require 3D spatial grounding and camera pose tracking. Second, action transcripts can introduce errors: on tasks requiring ordinal placement matching (e.g., VideoPlaceOrder), the ledger reaches $64.0\%$ against $100\%$ for oracle grounding, as temporal event semantics cannot be verified via geometry alone.

Despite these limitations, externalizing persistent world state into an explicit ledger provides clear advantages. Transparent object records make high-level reasoning interpretable and allow direct human intervention, such as updating object properties without modifying policy weights. Future work will focus on 3D mobile manipulation, real-world deployment, dynamic mid-episode object discovery, and further reducing supervisory fine-tuning as foundational VLMs evolve.

\section{Conclusion}
\label{sec:conclusion}

Where does each kind of memory belong? Short-term perceptual memory belongs inside the policy, as history-conditioned features, long-term spatial and event memory belongs outside the policy as an explicit spatio-temporal record, and a high-level planner chooses between them at runtime. \our{} realizes this split by pairing $\pi_{0.5}$ with a tracker-maintained ledger, a VLM-generated action transcript, and an LLM planner deciding at proprioception-confirmed step boundaries. On RoboMME, it is the only evaluated method competitive on all four suites and reaches the highest average success among them, $64.3\%$.

\begin{ack}

This work was partially funded by Office of Naval Research grant N00014-25-1-2322.

\end{ack}

\bibliographystyle{plainnat}
\bibliography{biblio}


\appendix

\section{The sixteen RoboMME tasks}
\label{app:tasks}
\begin{figure}[H]
\centering
\includegraphics[width=\textwidth]{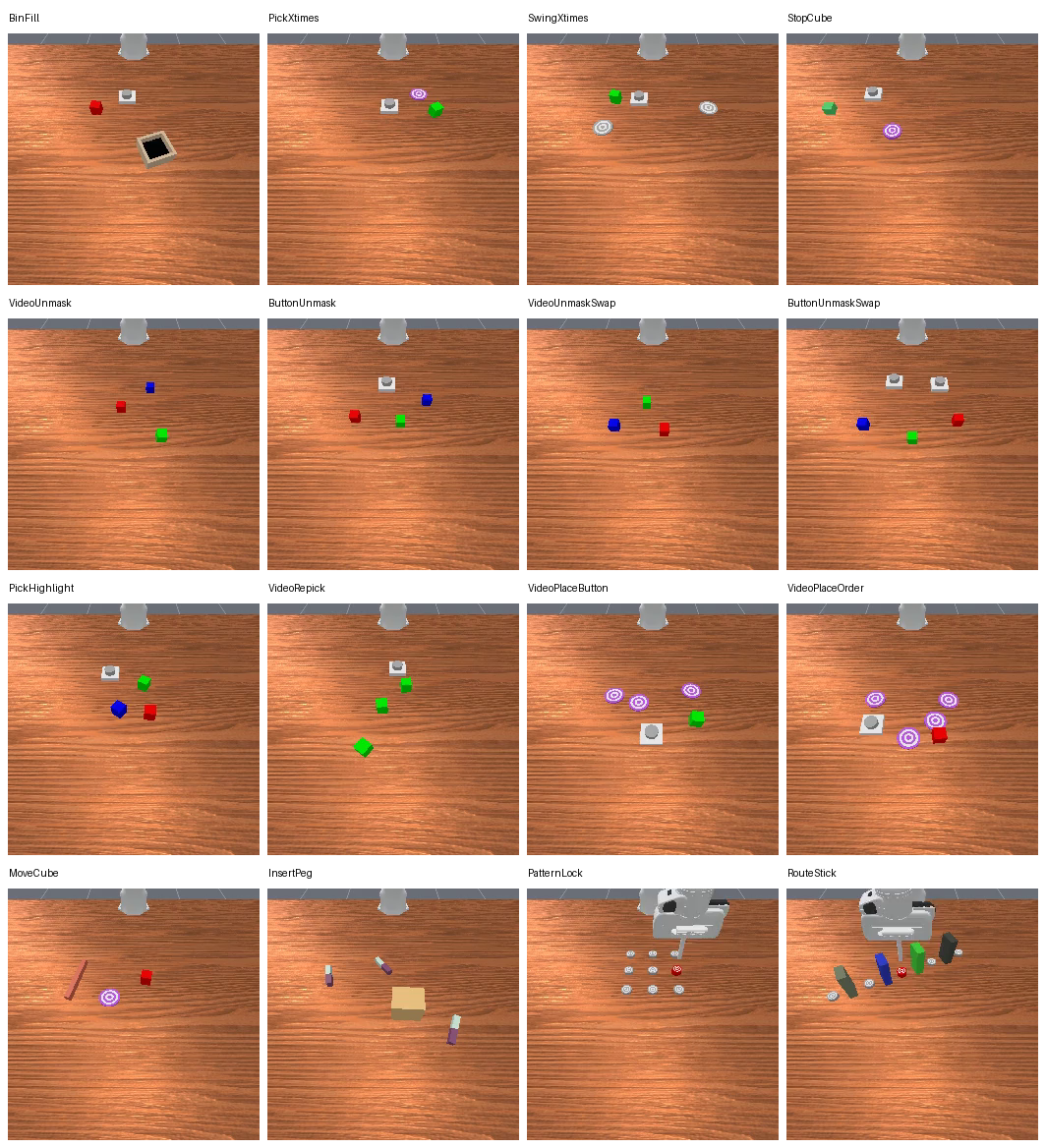}
\caption{First frame of one test episode of each RoboMME task, from our evaluation runs (front camera, $256 \times 256$, the resolution the policy and the tracker see). Rows are the four suites: Counting, Permanence, Reference, Imitation. Table~\ref{tab:instructions} gives the instruction of each task.}
\label{fig:tasks}
\end{figure}

Table~\ref{tab:instructions} lists one instruction per task; Figure~\ref{fig:tasks} shows a first frame of each.

\begin{table}[H]
\centering
\caption{One instruction instance per task, as given to the planner and to the low-level policy; object colours, counts, and targets vary across episodes. Tasks that begin with ``watch the video carefully'' start with a demonstration video; the others start from a single observed frame.}
\label{tab:instructions}
\footnotesize
\begin{tabular}{@{}lp{0.78\textwidth}@{}}
\toprule
Task & Instruction \\
\midrule
BinFill & put one red cube into the bin, then press the button to stop \\
PickXtimes & pick up the green cube and place it on the target, repeating this action three times, then press the button to stop \\
SwingXtimes & pick up the green cube, move it to the top of the right-side target, then move it to the top of the left-side target, repeating this back-and-forth motion two times, finally press the button to stop \\
StopCube & press the button to stop the cube just as it reaches the target for the fourth time \\
\midrule
VideoUnmask & watch the video carefully, then pick up the container hiding the green cube \\
ButtonUnmask & first press the button, then pick up the container hiding the red cube \\
VideoUnmaskSwap & watch the video carefully, then pick up the container hiding the green cube, finally pick up another container hiding the blue cube \\
ButtonUnmaskSwap & first press both buttons on the table, then pick up the container hiding the blue cube, finally pick up another container hiding the green cube \\
\midrule
PickHighlight & first press the button, then pick up all cubes that have been highlighteted [sic] with white areas on the table \\
VideoRepick & watch the video carefully, then repeatedly pick up and put down the same block that was previously picked up for three times, finally put it down and press the button to stop \\
VideoPlaceButton & watch the video carefully, then place the green cube on the target right after the button was pressed \\
VideoPlaceOrder & watch the video carefully, then place the red cube on the first target it was previously placed on \\
\midrule
MoveCube & watch the video carefully, then move the cube to the target in the same manner as before \\
InsertPeg & watch the video carefully, then grasp the same end of the same peg you've picked before and insert it into the same side of the box \\
PatternLock & watch the video carefully, then use the stick attached to the robot to retrace the same pattern \\
RouteStick & watch the video carefully, then use the stick attached to the robot to navigate around the sticks on the table, following the same path \\
\bottomrule
\end{tabular}
\end{table}

\section{Training and inference details}
\label{app:details}
The low-level policy is a $\pi_{0.5}$ VLA (PaliGemma-2B backbone with a flow-matching action expert) fine-tuned in full from the public \texttt{pi05\_base} checkpoint with the standard flow-matching objective for $80$k steps at batch size $64$, under the subgoal dropout schedule of Sec.~\ref{sec:frame}. The tracker is SAM3. The VLM captioner is Qwen3-VL-30B and takes about one minute per demonstration on one A100. Seeding, planning, and occlusion queries use Claude Sonnet 5.

\paragraph{Planner calls.} The planner is queried $2.7$ times per episode on average, from once on StopCube, PatternLock, and RouteStick to $4.8$ times on PickXtimes.

\section{Seed variance}
\label{app:variance}
Table~\ref{tab:variance} gives the mean and standard deviation over the three evaluation seeds of each suite mean and of the 16-task average, for the rows of Table~\ref{tab:main} that we ran ourselves.

\begin{table}[H]
\centering
\caption{Mean $\pm$ standard deviation over the three evaluation seeds of the suite means and of the 16-task average (success \%).}
\label{tab:variance}
\small
\begin{tabular}{lccccc}
\toprule
Method & Counting & Permanence & Reference & Imitation & Avg \\
\midrule
FrameSamp & $69.2 \pm 2.9$ & $25.7 \pm 1.3$ & $37.3 \pm 3.2$ & $51.3 \pm 2.5$ & $45.9 \pm 0.7$ \\
MemER & $49.2 \pm 1.6$ & $56.2 \pm 0.6$ & $40.3 \pm 1.9$ & $27.3 \pm 2.0$ & $43.3 \pm 0.7$ \\
Tracker & $66.0 \pm 1.8$ & $87.2 \pm 5.1$ & $32.7 \pm 0.8$ & $47.2 \pm 0.8$ & $58.3 \pm 1.8$ \\
\our{} & $66.8 \pm 2.8$ & $86.7 \pm 3.5$ & $60.7 \pm 5.2$ & $43.0 \pm 2.2$ & $64.3 \pm 0.9$ \\
\bottomrule
\end{tabular}
\end{table}

\section{The ledger as the planner reads it}
\label{app:ledger}
Below is the complete message the planner received at its first decision in one VideoUnmaskSwap episode with a single-container instruction, followed by its reply, taken from the run logs. Three cubes were seeded from the first frame, the tracker followed them through the 169-frame demonstration, and all three were reported covered at tick 35, when the containers came down. The transcript is wrong: it describes the red cube being stacked on the green one. The stationary intervals allow for the recovery: between ticks 88 and 112 the covers over A and B exchanged places (A's second rest at $(117.6, 86.3)$ is within \texttt{near\_px} of B's first rest at $(119.0, 93.4)$, and B's second rest is at A's first), and between ticks 142 and 162 they exchanged back. The planner needs none of this reasoning to act, as a covered object's \texttt{position} is where its cover sits now, so it grounds the pick at B's current position.

{\scriptsize
\begin{verbatim}
MEMORY:
{"objects": [
  {"id": "A", "description": "red cube (small red cube)", "position": [142.5, 73.3],
   "visible": true, "covered_since": 35,
   "rests": [{"t": [1, 99],    "position": [142.5, 76.5]},
             {"t": [112, 141], "position": [117.6, 86.3]},
             {"t": [162, 170], "position": [142.5, 73.4]}]},
  {"id": "B", "description": "green cube (small green cube)", "position": [119.1, 93.4],
   "visible": true, "covered_since": 35,
   "rests": [{"t": [1, 88],    "position": [119.0, 93.4]},
             {"t": [111, 142], "position": [144.8, 79.9]},
             {"t": [162, 170], "position": [119.1, 93.3]}]},
  {"id": "C", "description": "blue cube (small blue cube)", "position": [120.4, 127.8],
   "visible": true, "covered_since": 35,
   "rests": [{"t": [1, 170],   "position": [120.4, 128.0]}]}],
 "near_px": 12.8,
 "demonstration": "The robot arm lifts the red cube (A) and places it on top of the
   green cube (B), then lifts both cubes together and sets them down next to the blue
   cube (C). Finally, the robot arm presses down on the stacked cubes (A and B) while
   they remain in place beside C.\n\nPress timing, answered separately: There is no
   pressing action observed in the video. All movements involve lifting and placing
   the cubes, but none of the objects are pressed down by the robot arm. Therefore, no
   object is pressed, and there is no sequence of placements immediately before or
   after a press."}

COMPLETED THIS EPISODE:
  none yet

YOUR REMAINING PLAN FROM THE PREVIOUS DECISION:
  none: this is the first decision

GRIPPER NOW: empty

Instruction: watch the video carefully, then pick up the container hiding the green cube

Resolve the referents from the memory and output the remaining steps.
\end{verbatim}
}

\noindent The planner's reply, parsed strictly (a sentence from the executor vocabulary, one referent, a coordinate slot):
{\scriptsize
\begin{verbatim}
referents: B (green cube's container, identified by description/position; covered_since
shows a container sits over it)

{"steps": [{"action": "pick", "obj": "B",
            "subgoal": "pick up the container at <> that hides the green cube"}]}
\end{verbatim}
}

\noindent The harness fills the slot with B's live coordinate and hands the sentence to the low-level policy; proprioception confirms the grasp at tick 257, and the second decision receives the same memory with the completed-steps block filled in and a gripper line stating that an object is held:
{\scriptsize
\begin{verbatim}
subgoal handed to the policy:   pick up the container at <93, 119> that hides the green cube
COMPLETED THIS EPISODE (decision 2):
  1. pick up the container at <> that hides the green cube (obj B), confirmed at t=257
\end{verbatim}
}
\noindent The planner replies with an empty step list, and the episode ends as a success. The message also carries an annotated current frame, in which each tracked object is marked with its letter, that is not reproduced here.


\end{document}